\documentclass[conference]{IEEEtran}

\usepackage{cite}
\usepackage{amsmath,amssymb,amsfonts}
\usepackage{algorithmic}
\usepackage[hyphens]{url}
\usepackage{graphicx}
\usepackage{hyperref}
\usepackage{textcomp}
\usepackage{xcolor}
\usepackage[utf8]{inputenc}
\usepackage{makecell}
\usepackage[normalem]{ulem}  
\usepackage{xspace}
\usepackage{siunitx}

\def\BibTeX{{\rm B\kern-.05em{\sc i\kern-.025em b}\kern-.08em
   T\kern-.1667em\lower.7ex\hbox{E}\kern-.125emX}}

\begin{document}

\title{Can we Estimate Truck Accident Risk from Telemetric Data using Machine Learning?}

\author{
\IEEEauthorblockN{Antoine Hébert\IEEEauthorrefmark{1},
                  Ian Marineau\IEEEauthorrefmark{4},
                  Gilles Gervais\IEEEauthorrefmark{4},
                  Tristan Glatard\IEEEauthorrefmark{1} and
                  Brigitte Jaumard\IEEEauthorrefmark{1}}
\IEEEauthorblockA{\IEEEauthorrefmark{1}Department of Computer Science and Software Engineering \\
                  Concordia University, Montréal, Québec, Canada}
\IEEEauthorblockA{\IEEEauthorrefmark{4}Groupe Robert, Boucherville, Québec, Canada}
} 

\maketitle

\begin{abstract}
Road accidents have a high societal cost that could be reduced through improved risk predictions using machine learning.
This study investigates whether telemetric data collected on long-distance trucks can be used to predict the risk of accidents associated with a driver. We use a dataset provided by a truck transportation company containing the driving data of 1,141 drivers for 18 months.
We evaluate two different machine learning approaches to perform this task. In the first approach, features are extracted from the time series data using the FRESH algorithm and then used to estimate the risk using Random Forests. In the second approach, we use a convolutional neural network to directly estimate the risk from the time series data. We find that neither approach is able to successfully estimate the risk of accidents on this dataset, in spite of many methodological attempts. We discuss the difficulties of using telemetric data for the estimation of the risk of accidents that could explain this negative result. 
\end{abstract}

\begin{IEEEkeywords}
Road Safety, Data Mining, Machine learning
\end{IEEEkeywords}


\section{Introduction}

Despite improvements in road safety, road accidents remain an important issue worldwide:  they lead to an estimated 1.35 million deaths and more than 20 million injuries every year, and are the leading cause of death for people aged between 5 and 29~\cite{road_safety_report, traffic_injuries}. Road accidents also represent a high economic cost for society. In Canada, the yearly economic cost of transport-related injuries is estimated to US\$3.2 billions~\cite{road_accident_cost_canada}.

Road accidents are an important issue for truck transportation companies. Each accident can cause driver injuries, truck repair costs and the loss of transported goods. 
The US Federal Motor Carrier Safety Administration (FMCSA) estimated at US\$148,279 the average cost of a truck crash for society~\cite{FMCSA_report}. 
To minimize road accidents, most truck transportation companies analyze accidents to understand their causes and how they might be prevented. Some companies also offer regular training to their drivers to promote safe driving. According to the FMCSA, 5.5\% of fatal truck crashes are caused by driver fatigue and could have been prevented~\cite{FMCSA}.

In the United States and Canada, it is now mandatory for motor carriers to equip their trucks with electronic logging devices (ELD) directly connected to the vehicle to track service hours~\cite{ELD_rules, canada-eld}. This is an opportunity for transportation companies to go beyond the compliance requirements and install telemetric systems to collect a variety of sensor data from the vehicle. Many such telemetric solutions are available on the market to improve truck fleet management by providing real-time information to fleet managers~\cite{isaac}. Telemetric systems produce huge amounts of data, generated by an ever-increasing number of sensors on the vehicle.

The availability of big amounts of telemetric data generated by vehicles is a great opportunity to try to predict accidents by characterizing dangerous driving behaviour. Indeed, it is likely that the style of driving greatly influences the risk of accidents.
In this study, we design a machine learning model using such telemetric data to estimate the risk of accident associated with a driver.

Telemetric data generated by vehicles is in the form of time series. During driving, vehicle sensors record various parameters at regular intervals and store them in the telemetric system. We will therefore design machine learning models which can provide a measure of the risk of accidents of a given driver by looking at times series containing the evolution of various parameters during its driving. If we define the risk of accidents as the probability that this driver has an accident, then estimating the risk of accident is equivalent to classifying examples as leading to an accident or not. Therefore, the problem is a time series classification one.

Road accident prediction has been studied, but never using this type of data. Most studies predict the risk of accidents at different points in time and space using characteristics of the road network and weather information. Instead, we are interested in predicting the risk of accidents for a given driver based on information about their driving. Such a model could help truck transportation companies identify drivers with riskier driving styles, and offer them additional safe driving trainings. It could also be useful to insurance companies.

The rest of this paper is organized as follows:
Section \ref{sec:relatedwork} presents the related work on road accident prediction and on time series classification,
Section \ref{sec:datasets} and
Section \ref{sec:method} present our datasets and model creation methods,
Section \ref{sec:results} presents experiments and results, and Section \ref{sec:discussion} discusses these results.
Conclusions are drawn in the last section.

\section{Related work} 
\label{sec:relatedwork}
\subsection{Road Accident Prediction}

Many studies consider road accident prediction and aim at predicting the risk of an accident at a given place and time.
These studies would for example predict which segments of a road are most dangerous\cite{Chang2005}, or what times and areas of a city are most dangerous\cite{QChen2016}. They usually use information about the road such as the average daily traffic or the road curvature, as well as weather information such as the temperature or the precipitation.

Early work on road accident prediction used classical statistical modelling, usually variants of Poisson Regression. In 2005, Chang\cite{Chang2005} compared an artificial neural network with a negative binomial regression for the prediction of the number of accidents on road segments of a Taiwanese freeway: it was the first work to show that machine learning methods could achieve better performances than classical statistical modelling for road accident prediction. Later studies performed road accident prediction with various machine learning algorithms, usually only focusing on a few roads~\cite{Chang2005b, Lin2015, Theofilatos2017}. More recently, other studies performed road accident prediction at a larger scale covering larger areas or predicting at a higher-resolution~\cite{QChen2016, Najjar2017, Yuan2018, Hebert2019}. These studies showed that weather and road characteristics influence the risk of accident, and that it is possible to successfully identify places and times where accident are much more likely to happen. Instead, our goal is to identify the accident risk associated with a particular truck driver, regardless of location or weather conditions.

\subsection{Time Series Classification}

The literature on time series classification is very diverse in terms of methods and models. We identify four broad classes of methods: feature-based, model-based, distance-based\cite{abanda2019review}, and representation based\cite{fawaz2019deep}. Feature-based methods first derive features from the time series data and then apply classical classification algorithms. 
The model-based approach is a generative approach that trains, for each class, a generative model learning the characteristics of the class. To predict the class of a new example, each model is asked how likely it is that this example belong to its class, and the predicted class is the class with the highest probability.
Distance-based methods define a relevant distance metric between two time series and then use a k-nearest neighbor classifier (k-NN) or a support vector machine (SVM). Finally, representation-based methods use deep neural networks to learn a representation of time series and classify accordingly.

The performance of different methods highly depends on the type of time series and  problems.
The distance-based approach and the use of elastic distance measures were historically the most popular approach~\cite{bagnall2015time}.
Dynamic time warping (DTW) is a commonly used distance measure. Many variants have been proposed but Lines and Bagnall~\cite{lines2015time} have shown that none of them is significantly better than DTW.
In 2016, Bagnall et al. \cite{bagnall2017great} compared the performances of different time series classification methods from the feature-based, model-based and distance-based approaches on the datasets of the UCR time series classification repository\cite{UCRArchive2018}. The best-performing algorithm was COTE~\cite{bagnall2015time}, an ensemble of classifiers applied on various time series transformations. COTE combines $11$ distance-based classifiers and $24$ feature-based ones.
The same year, COTE was improved with HIVE-COTE\cite{lines2016hive} which introduces two additional sets of classifiers and a hierarchical voting system improving the aggregation of the different classifier results. An important limitation of both COTE and HIVE-COTE is their very high computational requirements as they combine many classifiers and complex transformations with complexities as high as $O(n^2t^4)$ with $n$ the number of time series and $t$ their length. This limitation makes it impractical to use these algorithms with big datasets or long time series.

When using feature-based or distance-based methods, it is hard to know which distance or which features to use without expertise on the data used. In 2016, Christ et al.~\cite{christ2016distributed} introduced an algorithm called FRESH (FeatuRe Extraction based on Scalable Hypothesis test) that automatically selects relevant time series features for binary classification. The algorithm has three main steps. First, it computes many possible features from the time series, simple features such as the mean, the standard deviation or the kurtosis, but also more advanced features such as the number of peaks or the spectral centroid. Then, for each feature, it uses a statistical test to check if the feature is relevant to predict the class, and finally selects the best features using the Benjamini-Yekutieli procedure\cite{benjamini2001control}. The resulting features can then be used with any classical machine learning algorithm. The authors evaluate the performances of this method when combined with an Adaboost classifier on the UCR time series classification repository\cite{UCRArchive2018}. It achieves results comparable to the DTW algorithm, with a lower computational cost as FRESH scales linearly with the number of samples and the length of the time series.
In 2019, Fawaz et al.\cite{fawaz2019deep} evaluated the performances of representation-based methods and compared the performance of several deep neural networks. They found that a ResNet deep neural network competes with HIVE-COTE while being much more computationally efficient. More recently, Fawaz et al.\cite{fawaz2019inceptiontime} introduced a new deep neural network architecture for time series classification slightly outperforming HIVE-COTE on the UCR time series classification repository with a win/draw/loss of 40/6/39. This new architecture named InceptionTime was inspired by the Inception-v4 architecture~\cite{szegedy2017inception} used in computer vision.

In summary, time series classification made significant progress in recent years. HIVE-COTE offers state of the art performances but has impractical computational cost. For big datasets, deep neural networks or the FRESH algorithm coupled with classical machine learning seems to be the two most promising approaches. We will use both approaches to build our models.

\section{Datasets}
\label{sec:datasets}

The datasets used in this study were collected by Groupe Robert Inc, a transportation company based in Quebec, Canada.
For many years, Groupe Robert Inc. has been monitoring road accidents and infractions involving their truck drivers to better understand how to reduce the number of accidents.
In 2017, it equipped its truck fleet with a telemetric system collecting most of the data generated by vehicle sensors during driving.

We used two datasets provided by the company: (1) the data from the sensors of the vehicle collected using the telemetric system onboard the trucks, and (2) the list of accidents involving drivers of the company, extracted from the records of the company. These datasets contain data collected for 18 months between February 2018 and June 2019.

The telemetric system records the values measured by the vehicle sensors whenever the engine is on. Different sensors are recorded at different time intervals, every half a second, every second, every 10 seconds or every minute. The values measured by the sensors are collected on the CAN BUS of the vehicle using the Society of Automotive Engineers J1939 communication protocol. This protocol defines identifiers for each sensor on the vehicle (see Table~\ref{table:sensors-list}). We have not used 24 other recorded parameters which we identified as not relevant for our study in agreement with the domain experts at Groupe Robert Inc. 

The truck fleet of the company is not homogeneous, it is composed of different types of trucks used for different transportation needs. The company identifies 3 different types of trucks: long-distance trucks, short-distance trucks, and specialized trucks like container and bulk trucks. These trucks are not equipped with the same sensors and follow different driving patterns. In our first model, we will focus on long-distance trucks, as it is likely that the other classes will require different models.

\begin{table}[htbp]
   \caption{Some of the parameters collected}
   \begin{center}
   \resizebox{\columnwidth}{!}{%
   \begin{tabular}{|r|l|}
      \hline
      Sensor identifier & Description \\
      \hline
      Acc\_Lat & Acceleration on the lateral axis\\
      Acc\_Long & Acceleration on the longitudinal axis \\
      Acc\_Long\_WBVS & \makecell[l]{Acceleration on the longitudinal axis \\ as measured on the wheels} \\
      Acc\_Vert & Acceleration on the vertical axis \\
      AccelPedalPos1 & Use of the acceleration pedal\\
      ActualEngPercentTorque & Engine torque in percentage \\
      ActualEnginePower & Engine power\\
      ActualEngineTorque & Engine torque \\
      ActualRetarderPercentTorque & Retarder torque in percentage \\
      AmbientAirTemp & Ambient air temperature\\
      BarometricPress & Barometric pressure\\
      BrakeSwitch & Status of brake switch \\
      CruiseCtrlActive & Status of cruise control \\
      EcoMode & Status of economy mode \\
      EngCoolantTemp & Temperature of engine coolant \\
      EngFuelRate & Fuel rate\\
      EngReferenceTorque & Engine reference torque\\
      EngSpeed & Engine rotation speed \\
      EngTurboBoostPress\_PSI & Engine turbocharger boost pressure \\
      EstEngPrsticLossesPercentTorque & \makecell[l]{Estimated torque loss \\ due to engine parasitics} \\
      NominalFrictionPercentTorque & Nominal friction torque in percentage \\
      Top\_Gear\_State & Whether the top gear is used \\
      WheelBasedVehicleSpeed & \makecell[l]{Vehicle speed as measured on \\ the wheels}\\
      gps\_Altitude & GPS altitude\\
      gps\_Lat & GPS latitude \\
      gps\_Long & GPS longitude \\
      gps\_Speed & Vehicle speed as reported by the GPS \\
      \hline
   \end{tabular}}
   \label{table:sensors-list}
   \end{center}
\end{table}

The company keeps track of all accidents involving their trucks, amounting to 1,434 accidents during the study period. For each accident, the date of the accident, the identifier of the driver, and the type of accident are recorded. Table~\ref{table:acc-types-list} shows the 30 types of accidents that were identified. The four most frequent types of accidents are types 1, 2, 3 and 4, representing 61\% of accidents. They correspond to non-severe accidents occurring mostly during maneuvers.

\begin{table}[htbp]
    \caption{Types of accident}
    \begin{center}
   \resizebox{\columnwidth}{!}{%
    \begin{tabular}{|r|r|l|}
        \hline
        ID & Proportion & Description \\
        \hline
        1 & 26\% & Accident while driving backwards \\
        2 & 17.6\% & Hit  a  stationary  object  (except  wall) \\
        3 & 6.6\% & Accident while changing  dock \\
        4 & 11\% & Hit  a  stationary  vehicle \\
        5 & 2.2\% & Hit  an  animal \\
        6 & 4.7\% & Rear collision \\
        7 & 1.5\% & Damaged  equipment  during  loading\\
        8 & 4.9\% & Miscellaneous \\
        9 & 2.3\% & Hit a cable \\
        10 & 3.5\% & Rubbing \\
        11 & 1\% & \makecell[l]{Accident while turning right at intersection because \\ a third party was overtaking on the right}\\
        12 & 0.2\% & Accident while going straight through the intersection \\
        13 & 3\% & Loss of control \\
        14 & 2\% & Accident or fined because the truck cut off \\
        15 & 1.9\% & Accident caused by trailer not properly coupled with truck \\
        16 & 0.9\% & Truck stuck (in snow for example), towing necessary \\
        17 & 2.1\% & Hit a wall or building \\
        18 & 0.7\% & Mechanical Breakdown \\
        19 & 1.7\% & Fined because of leaking truck \\
        20 & 1.2\% & Improper maneuvering in tight turns \\
        21 & 0.3\% & \makecell[l]{Fined because of improper snow clearance of the truck \\ (for example ice remaining on the truck roof)} \\
        22 & 0.5\% & Accident caused by vehicle wheel ignition \\
        23 & 0.7\% & Hit a bridge \\
        24 & 1.8\% & Equipment damaged during unloading \\
        25 & 0.6\% & Cargo \\
        26 & 0.1\% & Vehicle wheel loss \\
        27 & 0.3\% & \makecell[l]{Accident while turning left at intersection because  \\ a third party was overtaking on the left} \\
        28 & 0.1\% & Truck cargo theft \\
        29 & 0.7\% & Truck cargo fell out of the truck \\
        30 & 0.1\% & Equipment damaged without reported accident \\ 
        \hline
    \end{tabular}
    }
    \label{table:acc-types-list}
    \end{center}
\end{table}

\begin{figure*}[htbp]
\centerline{\includegraphics[width=\linewidth, keepaspectratio]{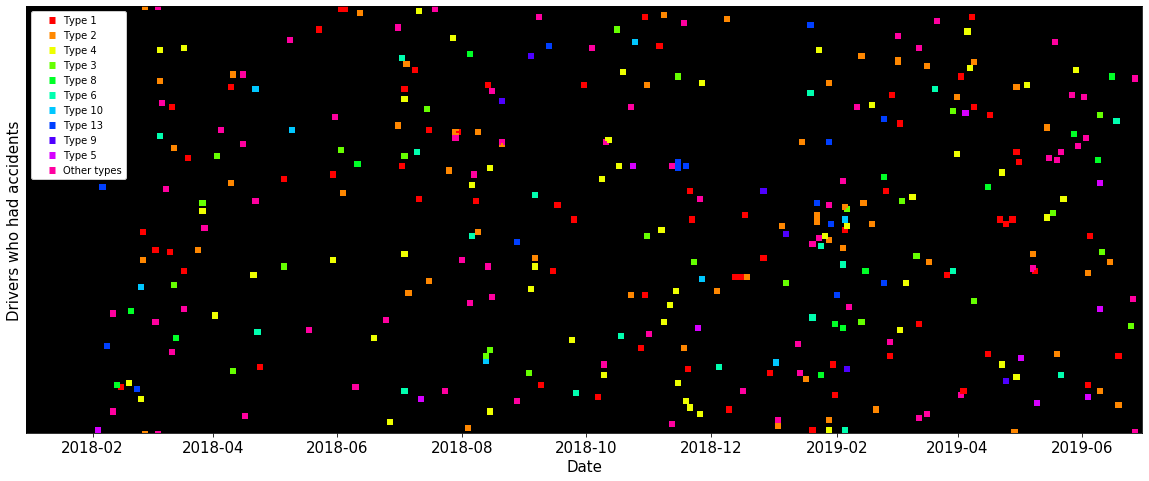}}
\caption{Visualization of the distribution of accidents in time}
\label{fig:acc_viz}
\end{figure*}

Figure~\ref{fig:acc_viz}  presents the distribution of accidents in time and across drivers. Each row corresponds to a different driver and the x-axis represents time. Each colored square correspond to an accident and the color of the square correspond to the type of accident. Only the 48\% of drivers who had an accident during the study period are included in this visualization. We observe that most drivers who had an accident had more than one during the study period.

\section{Method}
\label{sec:method}

\subsection{Data preprocessing}
The data obtained from Groupe Robert Inc required formatting to be usable for model training.
The data was initially in the form of 14 million files with each file containing the data collected on one truck during a driving period lasting between less than a minute to an hour. These files were in a proprietary format used by the telemetric system. Two MS Windows utilities were provided to convert a file from a proprietary format to another proprietary format and then to the CSV format. In addition, a separate CSV file identified which truck each driver was driving at different times. We used custom Python scripts and a virtual RAM drive to efficiently convert each file to the Apache Parquet format, using the provided utility to read files. The Apache Parquet format is a format from the Apache Hadoop ecosystem providing efficient data compression. This conversion was a data-intensive process that took several days. Once all files were converted to the Apache Parquet format, we identified the driver corresponding to each file and merged files corresponding to contiguous driving periods by the same driver on the same truck. As a result, we obtained 3.2 million Parquet files representing 890 GB of data.

We were informed that some of the accelerometer sensors might not be properly configured, and that the reported acceleration on the lateral axis, on the longitudinal axis and on the vertical axis might be permuted and in the wrong direction. We attempted to fix these issues by permuting and changing the sign of these parameters so that the acceleration on the longitudinal axis is positively correlated to the acceleration on the longitudinal axis as measured on the wheels for each truck and each month. This correction is not perfect since the accelerometers have been reconfigured at different dates for each truck and not necessarily at the beginning of the month.

\subsection{Instance creation}
\label{subsec:instance-creation}
\begin{figure}[htbp]
\centerline{\includegraphics[width=\columnwidth, keepaspectratio]{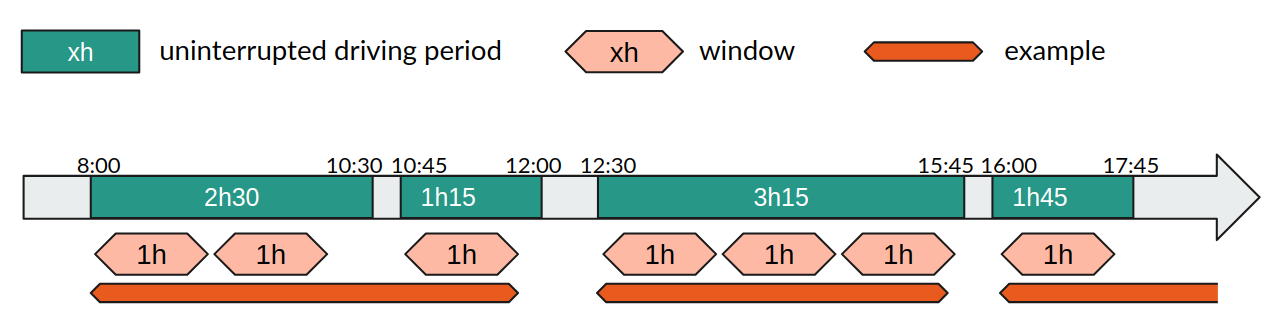}}
\caption{Illustration of example creation from raw data with 1-hour windows and 3 windows per example.}
\label{fig:example-extraction}
\end{figure}

Trucks make frequent stops which results in gaps in the time series. To alleviate this problem, we extracted non-overlapping windows of continuous driving from the raw data (Figure~\ref{fig:example-extraction}). After a few trials, we chose a window size of one hour, meaning that 3 windows of data could be extracted from a trip with a duration between 3 and 4 hours. A smaller window size would discard low-frequency patterns, while a too long window size would make it necessary to discard more data since driving periods shorter than the window size cannot be used. Since one hour of driving might not be enough to access the driving style of a driver, we aggregated 60 sequential but not necessarily contiguous windows to form each example. Therefore, our machine learning models look at 60 hours of driving to estimate the risk of accident. 

When performing statistical learning, we need to assume that examples are independent and identically distributed. In this study, we use the data from one driver to generate several examples, which means that examples are not all independent from each other. There is probably some correlation between examples corresponding to the same driver. This could affect learning, but it allows us to extract a reasonable number of examples from the limited data available. In the next subsection, we will show how we defined our test sets carefully so that they remain valid despite examples not being independent.

\begin{figure*}[htbp]
\centerline{\includegraphics[width=\linewidth, keepaspectratio]{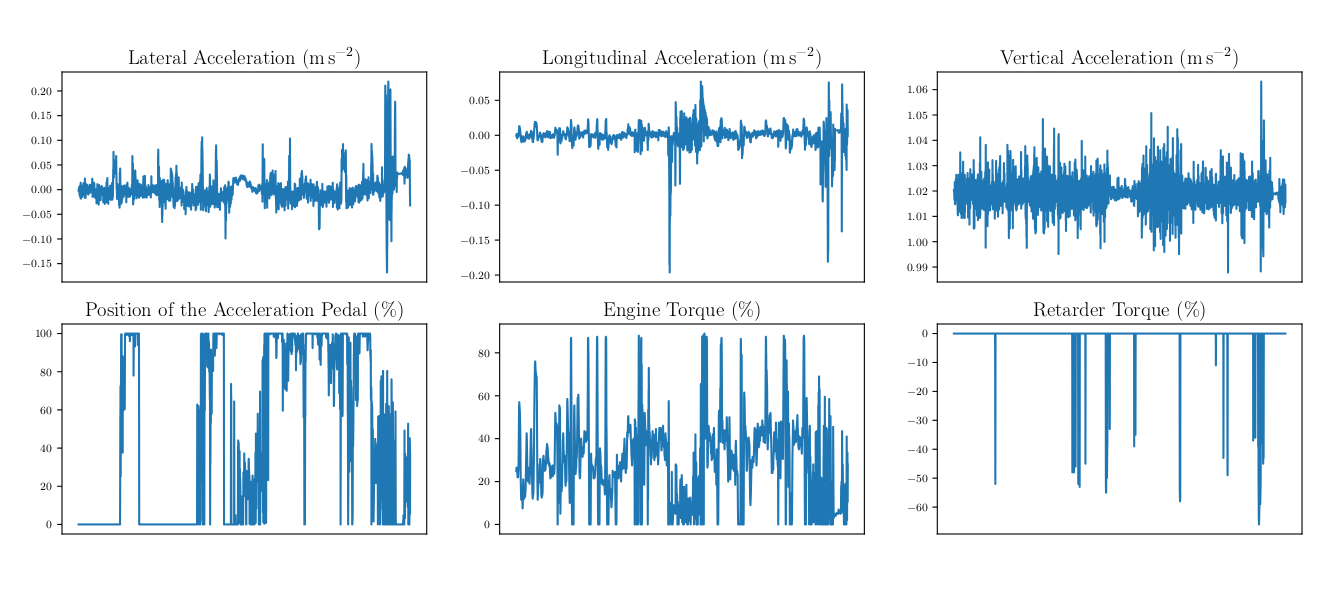}}
\caption{Window of one hour of driving data}
\label{fig:example-window-data}
\end{figure*}

As presented in Section \ref{sec:datasets}, a total of 51 parameters are recorded during driving, 24 of these parameters were identified as non-relevant by the domain experts from the company. We experimented with using various subsets of the 27 parameters left and found that the best results on the validation set were obtained when using only 6 parameters: the acceleration in the three dimensions, the position of the accelerator pedal, the engine torque and the retarder torque. The acceleration parameters and the engine torque were recorded every half a second while the other parameters were recorded every second. We downsampled the acceleration parameters and the engine torque to obtain the same sampling frequency for all parameters and reduce the computational requirement of further processings. Figure~\ref{fig:example-window-data} presents an example of the data corresponding to a one hour window. 

\subsection{Labeling}
\label{subsec:labeling}

Our goal was to obtain a model to estimate the accident risk. To train such a model in a supervised way, we needed for each example a ``ground truth" value of the risk of accidents. We used our second dataset, containing the list of accidents, to evaluate the accident risk associated with each example. By defining the risk of accident as the probability of having an accident, a model estimating the accident risk can be considered as a binary classification model. Driving data generated by a driver who had an accident belongs to the positive class, while data generated by a driver who did not belongs to the negative class. By training the model to classify driving data in this way, we obtained a model estimating the probability that new driving data belongs to the positive class, this probability is the accident risk according to our definition.

More precisely, we considered as positive the examples generated by a driver who had an accident in the year following the date of the example. We decided not to consider as positive the examples that followed an accident because we assumed that drivers might adjust their driving after they have an accident. We used a duration of one year because accidents are rare, and an incautious driving will not result in an accident right away. We experimented with shorter durations ranging from a week to a year.

As explained in Section~\ref{sec:datasets}, there are different types of accidents in the dataset. It is likely that some of these accident types are not related to the driving data, for example drivers are probably not responsible for accidents of type 5 when the truck hit an animal. 
Therefore, we decided to ignore some accident types, based on how well they are predicted on the validation set. We only used the accident types 1, 2, 7, 8, 9, 11, 15, 16, 17, 22, 23.

\subsection{Creation of training and test sets}
\label{subsec:creation-train-test-sets}
As mentioned previously, there is a risk of shared information between the training and test sets due to the way we create examples. This has consequences on how to correctly split the examples into a training set and a testing set for performance evaluation.

If we simply take a random sample of examples to create the test set, we will be measuring the ability of the model to recognise drivers, and not its ability to measure the accident risk of a new driver. Indeed, examples from the same drivers would be present in both the training and the testing set. In addition, most of the examples generated from the data of one driver have the same label: if the driver never had an accident during the study period, then all the examples will be negative; if the driver had an accident toward the end of the study period, then almost all their examples will be positive. Therefore, the model could correctly classify an example simply by recognizing the driver and retrieving from the training data whether this driver had an accident after the example occurred.

We split the training and test sets by driver rather than by example, to make sure that we evaluate the ability of the model to estimate the accident risk on unknown drivers. In addition, to ensure that the test set is a representative sample of the examples, we performed a stratified split: we ensured that the percentage of positive examples in the test set is approximately the same as the percentage of positive examples in the data. We used approximately 30\% of the examples to create the test set. 

We did not use the test set for the tuning of preprocessing and for model selection. Instead, we used validation sets created from the training data with the same procedure as for the test set, i.e., by making a stratified split by driver. We performed early tuning with one validation set containing 30\% of the training examples. We noticed that reported performance metrics could significantly change depending on which random validation set was used, so we later used k-fold cross-validation to obtain a more stable estimation of performances. Like for the test set, we made sure that examples corresponding to one driver were either in the training set or the validation set and that the proportion of positive examples in the validation set was representative of the proportion of positive examples in the dataset.

\subsection{Feature-based approach}
We built a first model using a feature-based approach and the FRESH algorithm~\cite{christ2016distributed} for feature extraction and selection. Indeed, as discussed in Section~\ref{sec:relatedwork}, the FRESH algorithm seemed a promising approach for time-series classification when using big datasets.

We used the TSFRESH library~\cite{christ2018time} (version 0.14.1), a Python library implementing the FRESH algorithm to extract features from time series and select the most promising ones. The extraction of these features for all the examples was a long process that took several days. To speed up the process, we excluded features labeled by the library as having a high computational cost. A total of $4,488$ features were extracted for each example, and $1,728$ of them have been considered relevant by the FRESH algorithm. We used the Random Forest algorithm~\cite{Breiman2001} to perform classification based on these features. Hyperparameters of the Random Forest algorithm were tuned using 5-fold cross-validation.

\subsection{Representation-based approach}
As discussed in Section~\ref{sec:relatedwork}, deep neural networks have obtained state-of-the-art performances on some TSC datasets and offer a much lower computational cost than competing methods. 

We started with the neural network architecture which obtained the best average performance in \cite{fawaz2019deep}: a ResNet neural network\cite{he2016deep} adapted for TSC.
This architecture is composed of 3 residual blocks followed by a global averaging pooling averaging feature maps over time and a final fully-connected layer. Each residual block is composed of 3 convolutional layers using batch normalization and a residual connection adding the input of the block to the pre-activation of the last layer. This residual connection is the main characteristic of this architecture and gave it its name which stands for Residual Network. Fawaz et al. provide an implementation of this neural network using TensorFlow, which we reimplemented in PyTorch~\cite{NEURIPS2019_9015} for convenience.

The original architecture takes as input matrices of dimension $(C, L)$ with $C$ the number of channels and $L$ the the length of the time series. We adapted the neural network to be able to use tensors of dimension $(N, C, L)$ with $N$ the number of windows. As indicated in \ref{subsec:instance-creation}, we used $N = 60$ windows for each examples.
We adapted the architecture by removing the last layer and applying the rest of the neural network to each window. We added a head combining extracted features. We initially used a few fully-connected layers to form the head, but later found that a global average over windows followed by one final fully-connected layer seemed to perform best.

Our initial adapted ResNet obtained very bad performances on the validation sets. We made a lot of changes to the neural network architecture and its training procedure to obtain better performances.

We quickly noticed that our model was subject to overfitting, indeed, while it obtained very good results on the training set, results on the validation sets were very bad. We therefore added spatial dropout~\cite{tompson2015efficient} after each convolutional layer to regularize the model. Spatial dropout consists in randomly dropping out the activation of some feature maps during training. With convolutional layers, it is recommended to use spatial dropout instead of regular dropout, indeed neurons from the same feature map are usually correlated and randomly dropping neurons independently does not affect much the learning process. We used a high dropout rate for all layers in order to regularize our model as much as possible. We found using automatic hyperparameter tuning that a dropout rate of 57\% seemed to perform best on the validation sets.

Even after adding heavy dropout to the neural network, and reducing the number of feature maps, the neural network was still overfitting the training data. In order to further reduce its capacity, we tried reducing its depth. We found that the neural network was performing best on the validation set with only one residual block. This is surprising because deeper networks trained for less epochs usually generalize better than shallower network trained for longer. With such a shallow network, one could wonder if the residual connection is still useful, after experimenting without we found it was indeed not useful. We also removed zero-padding  which became no longer necessary.

To further reduce the capacity of the model, we tried making use of strided convolutions. By using a convolutional layer with a stride greater than 1, the following convolutional layer can achieve the same receptive field with a smaller kernel. We found better results when using a stride of 2 for the first two convolutional layers while adapting the kernel sizes accordingly.

We experimented with different activation functions. The Exponential Linear Unit (ELU)\cite{elu} seemed to perform best, so we replaced the ReLU activations initially used by ELU activations.

It can be challenging to find the right set of hyperparameters for which a neural network will learn successfully. The common practice is to start with the configuration of hyperparameters used by another study on a related problem. It was not possible in this study since to the best of our knowledge, there are  no other studies making use of telemetric driving data for accident prediction.
To help with the search of a good configuration of hyper-parameters, we made use of automatic hyper-parameter tuning. Thanks to the limited size of our dataset and of our model, it was possible to try many different configurations. The following hyperparameters were automatically tuned: the amount of weight decay, the dropout rate, the kernel size of the three convolutional layers and the number of feature maps. We found that the amount of weight decay did not seem to matter, this might be because the use of batch normalization changes the effect of weight decay~\cite{weight_decay_batchnorm}. For other hyperparameters, we obtained the following values: 57\% for the dropout rate, 31, 8 and 4 for the kernel sizes of the first, second and third convolutional layers and 10 for the number of feature maps.

To train the neural network, we used the Adam optimization algorithm~\cite{adam} with a small amount of weight decay. We used the corrected implementation of Adam with weight decay~\cite{adamw}. We used a batch size of 32. To find a good learning rate, we used the method presented in \cite{lrfinder}, and we obtained a learning rate of \SI{1.1e-1}{}. To determine for how many epochs to train the model, we used early stopping: we evaluated the performances of the model on the validation set after each epoch and stopped training when the performance did not improve for $3$ epochs in a row. Finally, we used a focal loss~\cite{focalloss} instead of the usual cross-entropy loss. This loss is designed to help with data imbalance and we found that it improved our results.

\section{Experiments and Results}
\label{sec:results}

To measure the performance of our models, we used mainly the area under the Receiver Operating Characteristic (ROC) curve. The ROC curve shows the evolution of the True Positive Rate (TPR) as a function of the False Positive Rate (FPR) when varying the threshold used by the model to classify examples. The TPR is the proportion of examples identified as positives among actual positives, and the FPR is the proportion of examples identified as positives among actual negatives. The area under the ROC curve corresponds to the probability that the model will rank a randomly chosen positive example higher than a randomly chosen negative one, so we believe it is appropriate to evaluate a risk estimation model. 

As indicated in the previous section, for both approaches, we used k-fold cross-validation for model selection. We decided to report results on both the validation sets and the test set. For the validation results we report the average of the results of the different models obtained with different splits of the training data. To obtain the average ROC curves, we average the True Positive Rate for each False Positive Rate. For the test results, with the feature-based approach, we simply retrain a model using the whole training dataset before evaluating it on the test set. With the representation-based approach, the validation set is not only used for model selection but also for early-stopping, so we cannot retrain a single model using the whole training data. Instead, we report the average results on the test set of the models obtained with different splits. We cannot simply select the model with the best validation results among the models trained with different splits, because the performances of the model on the validation set do not reflect its performances on the test set.

With the feature-based approach, we obtained an average area under the ROC curve of $58$\% on the validation sets, which correspond to performances slightly better than those of a random classifier. But on the test set we obtained an area under the ROC curve of $43$\% only. 

With the representation-based approach, we were able to obtain better results on the validation sets with an average area under the ROC curve of $65$\%. On the test set however, results are the same as with the feature-based approach with an area under the ROC curve of $43$\%.

With both approaches, we noticed a high variation of performances measured using the validation set across the different splits of the k-fold cross-validation. The standard deviation of the area under the ROC curves was $9$\% with the feature-based approach and $7$\% with the representation-based one.

\begin{figure}[htbp]
\centerline{\includegraphics[width=\columnwidth, keepaspectratio]{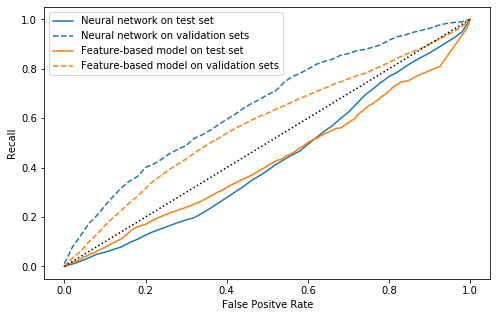}}
\caption{ROC curves of the feature-based model and of the neural network on the test set and on the validation sets}
\label{fig:roc1}
\end{figure}

Figure~\ref{fig:roc1} presents the ROC curves on the test set and on the validation sets obtained with both models.

\begin{figure}[htbp]
\centerline{\includegraphics[width=\columnwidth, keepaspectratio]{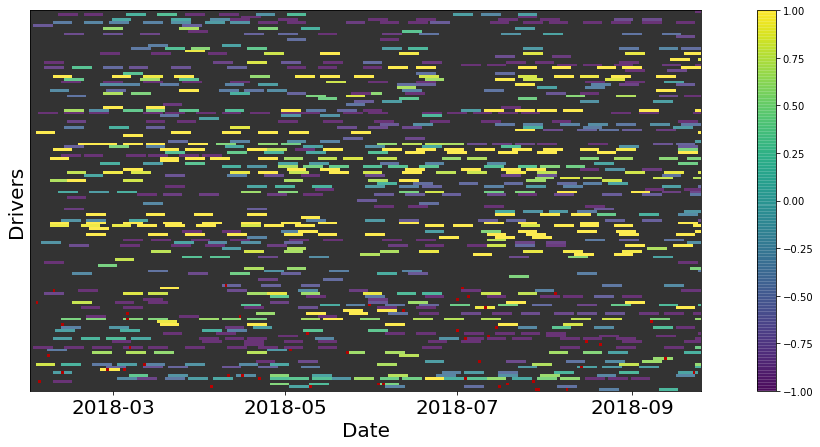}}
\caption{Visualization of the risk of accidents estimated by the representation-based model}
\label{fig:pred_viz}
\end{figure}

Figure \ref{fig:pred_viz} shows a visualization of the risk of accidents estimated by the neural network model on examples of the test set. Each row corresponds to a different driver and the x-axis represents time. Each colored rectangle correspond to an example, and its color represents the prediction of the model. 
On this figure, we observe that the model usually estimates the same risk of accident for different examples corresponding to the same driver at different times during the study period. This is interesting because the model has no knowledge of which driver an example correspond to.

\section{Discussion}
\label{sec:discussion}

With performances on the test set worse than those of a random classifier, we cannot say that we were able to estimate the risk of accident accurately in this study. In this section, we  discuss the reasons that could explain those results.

Road accidents are caused by a combination of many factors: how the driver drives, but also on which road they drive and under which weather and traffic conditions. In this study, we only use data describing the driving: we expected that by looking at whether a full-time driver had an accident during a long study period of 18 months, these other factors would average out. That is to say that during the study period, drivers would have met all kind of driving conditions and that on average drivers with accidents would show a different driving style. Given the results we obtain, the study period or the number of drivers might not be long and high enough for this to happen.

In addition, the driving style of a given driver is likely to significantly vary over time. Indeed, quality of driving is likely to be affected by the hour of the day and fatigue. This means that a driver who had an accident during the study period because they were particularly tired on that day might not necessarily show dangerous driving patterns during the rest of the study period, and our labelling method would result in misclassified training examples. This would suggest to label examples as dangerous only when they occur during the few days before an accident. However, it might also happen that a driver always drives dangerously, but because accidents are very rare only has one or even no accident during the study period. This would also result in many misclassified training examples. We can also imagine cases where very careful drivers are involved in an accident due to other factors such as bad weather conditions or bad behaviors of other users of the road. These problems result in a very noisy labelling of examples. Machine learning can work with noisy labels as long as the majority of examples are correctly classified, but it requires more examples or a high inductive bias. It could be interesting to experiment with semi-supervised learning methods and different labeling methods to see if they would help to deal with these issues. 

Our telemetric data might not be able to describe driving behaviors accurately enough to estimate the risk of accidents. For example,
information about the use of the steering wheel is only available through the lateral accelerometer, as there is no sensor on the steering wheel.
An important part of driving is the observation of everything that is happening outside the vehicle. A good driver not only drives carefully and maneuvers smoothly, but also consistently and accurately monitors everything happening on the road. The sensors we have access to do not give information about this important part of the driving activity. This part of driving is especially important for our dataset, because as we have seen in Section~\ref{sec:datasets}, most accidents do not happen on the road, but at slow speed during maneuvers. It could be interesting to add sensors in trucks to collect data about visual checks performed by the driver. We believe that recent improvements in computer vision make it possible to use a camera aboard the vehicle to determine whether visual checks have been performed for example. 

The relation between telemetric data and the risk of accident is complex. A more dangerous driver is probably not simply a driver with an higher average speed, it is likely to be a driver for which the evolution of telemetric data in specific contexts follow different patterns than safer drivers. For example, we might be able to evaluate to what extent a driver anticipates turns by looking at the evolution of the speed before a turn. With the representation-based approach, this would mean that the transformations from the raw data to a useful representation are quite complex. For this reason, we think that a neural network capable of applying such a complex transformation would require many layers and maybe more powerful structures than only convolutional layers. For example, the use of attention might make sense for our task, as it would allow the neural network to focus on windows of the time series that are particularly useful to assess a driver's driving style. However, the limited size of our dataset does not make it possible to train such networks, and for this reason we experimented mostly with relatively small networks for this study.

Another difficulty that we face when using machine learning to predict rare events like road accidents is the data imbalance issue. Indeed, machine learning algorithms tend to focus on the majority class and fail to account for other classes. It is quite easy to deal with this issue by assigning a higher weight to examples of the minority class, or by resampling the dataset. However, data imbalance sometimes hides another issue which is harder to deal with: a too small sample of examples for one class. With too few examples from one class, it is harder for the algorithm to learn significant characteristics of the class to discriminate it. Accidents are rare, so most examples belong to the negative class. Combined with the limited size of our dataset, it makes it harder for the models to train without overfitting the positive examples of the training set.

Our results show that there is a big difference between the performances of the models on the validation sets and their performances on the test set. This might be because many hyperparameters were determined by looking at the performances on the validation sets. The validation set was used to determine how to create instances: the length of the windows and the number of window per example. It was also used to determine how to label examples, to choose which accidents are considered predictable and for how long before an accident the driving data is labelled as positive. Finally, it was also used to determine the list of sensors to use and the hyperparameters of the models. Some of these hyperparameter values might be indeed better in general, but some of them might be particularly better just for the limited training and validation datasets and artificially increase performances reported using the validation set. Because of the limited size of the dataset and the issue of noisy labels discussed earlier, the measure of the performances on a subset of data is probably not reliable enough to take decisions. Indeed, the standard deviations of the areas under the ROC curve obtained with different splits of the k-fold cross-validation are quite high ($7$\% and $9$\% respectively for the feature-based approach and the representation-based approach). Given these limitations with the validation procedure, it might have been best to rely more heavily on our intuition and understanding in order to take some of the model design decisions.

In Figure~\ref{fig:pred_viz}, we observed that the neural network model usually estimates the same risk of accidents for examples from the same driver at different dates. This suggests that the model bases its prediction on characteristics of driving that are invariant over time for a driver. This could be because the accident risk indeed does not change much over time for a driver, but it could also be simply because of the way the model is trained. During training, the model does not know that we want it to predict the risk of accident, it only has access to pairs of driving data and labels. As discussed in part \ref{subsec:creation-train-test-sets}, most examples from the same driver have the same label. Because of this, the most simple way to learn the mapping between driving data and labels might be to simply recognise drivers. Once the model has learned to recognise drivers from the training set, it can already achieve an almost perfect score on the training data. When presented an example from a new driver, such model would try to recognise the driving of a driver from the training set and output the accident risk of this driver. This behavior would lead to the kind of results we observe, most examples from the same new driver would look like the same driver form the training set and the model would therefore output the same accident risk.

In other words, this effect could be caused by the fact that most examples from the same driver in the training set have the same label and that it might be easier to identify the driver than to estimate the risk of an accident using the driving data. In order to force the model to learn to recognise safe driving as opposed to who his driving, we might need a higher number of different drivers in the training set. With more drivers, it would becomes more difficult for the model to learn what the driving data of each driver look like and become necessary for the model to start making links between the driving data of different drivers with the same labels. A different approach could also help to deal with this issue without requiring a higher number of different drivers. For example, we could frame the problem as a meta-learning problem for which each task consists in classifying driving data from one driver depending on whether it was followed by an accident or not. This would prevent the model from cheating by recognising the driver since each episode would contain only data from one driver. By using meta-learning, the meta-model could learn how to train a good accident risk estimator by putting together knowledge from different drivers. 

Our results show that using telemetric data to estimate the risk of accident of a particular driver is not easy, but it does not mean that it is impossible. It might require a bigger dataset and a community of researchers and machine-learning practitioners to find the right approaches and methods. For this reason, we think it would be useful to publish our dataset and make it accessible to anyone who wants to work on this problem. However, the publication of such a dataset raises important ethical issues, as the raw dataset contains personal information such as the GPS position and the work schedule of the drivers which cannot be published. Publishing a preprocessed version of the dataset would restrict the way one can frame the problem.

\section{Conclusion}
\label{sec:conclusion}
We can still not give a definite answer to the question: ``Can we estimate truck accident risk from telemetric data using machine learning?". In our study, with the dataset we had access to, it is unlikely to be possible. Indeed, we experimented with two different approaches and many different methods without success. It would be interesting to see if this task would become possible with larger datasets, including more drivers and with data from different sensors. We believe that the estimation of the risk of accidents of a driver based on its driving data remains a very difficult machine learning problem. Indeed, because of the many factors that determine the occurrence of an accident, the number of accidents does not seem to be a good surrogate for driving quality. It might be necessary to use a different approach to teach a machine learning model what safe driving looks like.

\section{Acknowledgement}
We warmly thank Isaac Instruments, the company designing and producing the telemetric solution installed on the truck fleet of Groupe Robert, for facilitating the transfer of the data and their interpretation.

We are also grateful to Compute Canada and its regional centers Calcul Qu\'ebec and WestGrid for providing access to the computing clusters used in this study.

\bibliographystyle{IEEEtran}
\IEEEtriggeratref{23}
\bibliography{IEEEabrv,biblio}

\end{document}